\pdfoutput=1

\documentclass[10pt,twocolumn,letterpaper]{article}

\usepackage[pagenumbers]{cvpr} 

\usepackage[accsupp]{axessibility}

\usepackage{graphicx}
\usepackage{amsmath}
\usepackage{amssymb}
\usepackage{booktabs}

\usepackage{graphbox}
\usepackage{paralist}

\usepackage{multirow}
\usepackage{adjustbox}
\usepackage{arydshln}
\usepackage{verbatim}

\usepackage[table, svgnames, dvipsnames]{xcolor}
\usepackage{makecell}
\usepackage{tabularx}
\usepackage[flushleft]{threeparttable}
\usepackage{algorithm}
\usepackage{setspace}
\usepackage[noend]{algpseudocode}
\usepackage{multirow}
\usepackage{color}
\usepackage[dvipsnames,svgnames]{xcolor}

\usepackage{pifont}
\newcommand{\cmark}{\ding{51}}%
\newcommand{\xmark}{\ding{55}}%

\usepackage[normalem]{ulem}

\definecolor{MyGreen}{cmyk}{100, 0, 100, 0}
\usepackage[pagebackref,breaklinks,colorlinks,citecolor=MyGreen]{hyperref}

\usepackage[capitalize]{cleveref}

\crefname{section}{Sec.}{Secs.}
\Crefname{section}{Section}{Sections}
\Crefname{table}{Table}{Tables}
\crefname{table}{Tab.}{Tabs.}

\newcommand{\figcspace}{\vspace{-2mm}}
\newcommand{\figspace}{\vspace{-4mm}}
\newcommand{\tabcspace}{\vspace{-2mm}}
\newcommand{\tabspace}{\vspace{-4mm}}

\def\cvprPaperID{****} 
\def\confName{CVPR}
\def\confYear{2022}

\begin{document}

\title{Attentive Fine-Grained Structured Sparsity for Image Restoration}

\author{
\vspace{0.5em} 
Junghun Oh$^{1}$ \ \ Heewon Kim$^{1}$ \ \  Seungjun Nah$^{1,3}$ \ \  Cheeun Hong$^{1}$ \ \  Jonghyun Choi$^{4}$ \ \  Kyoung Mu Lee$^{1,2}$\\
$^{1}$Dept. of ECE, ASRI, $^{2}$IPAI, Seoul National University \quad $^{3}$NVIDIA \quad $^{4}$Yonsei University\\
{\tt\small $^{1}$\{dh6dh, ghimhw, cheeun914, kyoungmu\}@snu.ac.kr, $^{3}$seungjun.nah@gmail.com, $^{4}$jc@yonsei.ac.kr}
}

\maketitle

\begin{abstract}

Image restoration tasks have witnessed great performance improvement in recent years by developing large deep models.
Despite the outstanding performance, the heavy computation demanded by the deep models has restricted the application of image restoration.
To lift the restriction, it is required to reduce the size of the networks while maintaining accuracy.
Recently, $N$:$M$ structured pruning has appeared as one of the effective and practical pruning approaches for making the model efficient with the accuracy constraint.
However, it fails to account for different computational complexities and performance requirements for different layers of an image restoration network. 
To further optimize the trade-off between the efficiency and the restoration accuracy, we propose a novel pruning method that determines the pruning ratio for $N$:$M$ structured sparsity at each layer.
Extensive experimental results on super-resolution and deblurring tasks demonstrate the efficacy of our method which outperforms previous pruning methods significantly.
PyTorch implementation for the proposed methods is available \href{https://github.com/JungHunOh/SLS_CVPR2022}{here}.
\end{abstract}

\vspace{-0.2cm}
\section{Introduction}
\label{sec:introduction}

\begin{figure}[!t]
    \newcommand{\wwp}{0.99\linewidth}
    \centering
    \includegraphics[width=\wwp]{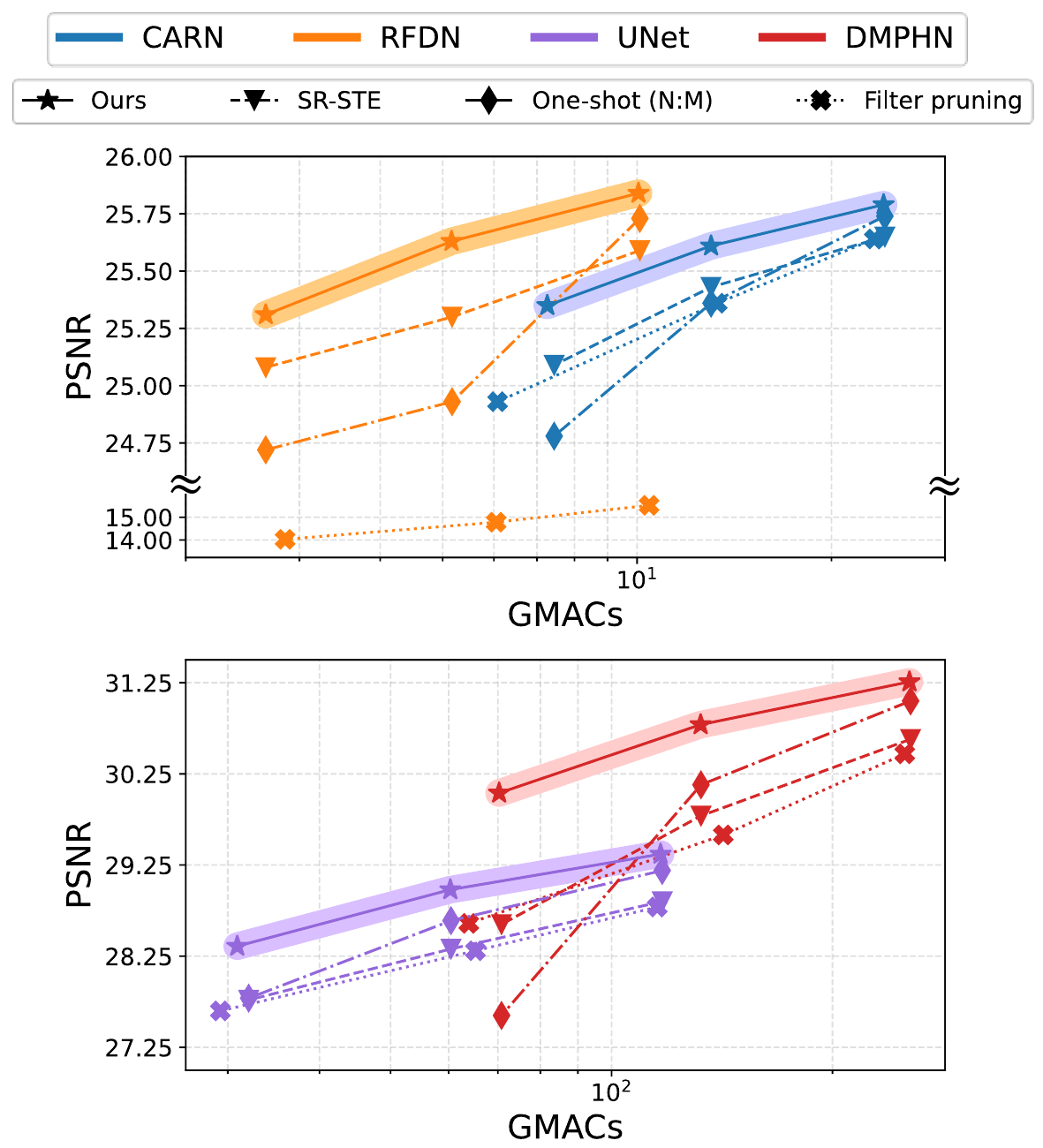}
    \figcspace
    \caption{Trade-off between image restoration performance (PSNR) vs computational costs (MACs) on super-resolution (top) and deblurring tasks (bottom).
    We compare our method to the magnitude-based filter pruning~\cite{li2017pruning} and the existing methods on $N$:$M$ sparsity (One-shot pruning~\cite{mishra2021accelerating} and SR-STE~\cite{zhou2021learning}).
    }
    \vspace{-0.5em}
    \figspace
    \label{fig:adaptive_sparsity_visualization}
\end{figure}

Advances in deep learning has brought success in image restoration tasks such as super-resolution~\cite{lim2017enhanced,son2021srwarp} and deblurring~\cite{Nah_2017_CVPR,Zhang_2018_CVPR_svrn,Zhang_2019_CVPR_DMPHN}.
Due to the heavy computational burden required by such methods, however, computing high-resolution images in practical applications has been challenging.
Network pruning is one of the most popular tools to alleviate the computational burden of neural networks by eliminating weights that are less critical to the accuracy.
It has shown remarkable efficacy in finding submodels for a better trade-off between accuracy and efficiency for image classification~\cite{he2017channel,liu2019metapruning,guo2020dmcp,lin2020hrank,he2020learning} and segmentation~\cite{you2019gate,he2021cap}.

Unstructured pruning~\cite{lecun1990optimal,hassibi1993optimal,han2015learning} aims to find and remove individual weights that have relatively less impact on model accuracy.
Still, accelerating the resulting models is difficult due to irregular sparsity patterns of weight tensors, considering the complex nature of parallelization on GPUs.
On the other hand, structured pruning removes predetermined structures (\eg, a filter in convolution layers~\cite{li2017pruning} or a channel of feature maps~\cite{he2017channel}) to enable the acceleration of pruned networks on GPUs.
However, we empirically find image restoration models to be often susceptible to substantial performance degradation from structured pruning.

Recently, $N$:$M$ fine-grained structured sparsity~\cite{mishra2021accelerating,hubara2021accelerated,zhou2021learning} has emerged as a better alternative, combining the strengths of both pruning methods: namely, fine-grained sparsity from unstructured pruning and hardware acceleration-ability from structured pruning.
$N$:$M$ structured sparsity enforces $N$ number of weights in each group of $M$ number of consecutive weights to have non-zero values.
Such sparsity constraint has the potential of hardware acceleration, where $2$:$4$ sparsity pattern has recently been supported in NVIDIA Ampere generation GPUs~\cite{mishra2021accelerating}.
However, training a network with $N$:$M$ sparsity has been proven to be difficult, which the existing works on $N$:$M$ sparsity have focused on improving~\cite{mishra2021accelerating,zhou2021learning}.
Such difficulty has prevented the direct application of the already developed pruning techniques, such as pruning with layer-wise varying pruning ratios~\cite{li2020eagleeye,liu2019metapruning}, which is known to be crucial to the performance of the pruned networks.
Particularly, we observed that several layers (\eg the last upsampling layer) in image restoration networks are very sensitive to pruning with respect to the performance.

Here, we propose a layer-wise $N$:$M$ sparsity search framework for efficient image restoration networks, named \textbf{S}earching for \textbf{L}ayer-wise $N$:$M$ structured \textbf{S}parsity (SLS).
In the prior arts~\cite{guo2020dmcp,lin2020hrank,he2017channel,liu2017learning,lin2019towards}, a filter or a channel is used as a unit of pruning but it is challenging to define the pruning unit in the case of $N$:$M$ sparsity.
To this end, we propose to consider the original weight tensor as the sum of sparse tensors whose configurations are determined by the magnitude of weights.
We use each of sparse tensor as the unit of pruning in our $N$:$M$ sparsity search problem.
To learn how many units to preserve, we propose a trainable score for each pruning unit, which is designed to ensure units with the lowest magnitude-based importance are removed first for better performance.

Furthermore, since image restoration tasks often have different computational constraints,  we present an adaptive inference method that uses several models trained by SLS with different efficiency.
The proposed adaptive inference technique determines which pruned model should be used at inference time depending on the restoration difficulty of an input image patch.
The adaptive inference method further improves the efficiency-accuracy trade-off and enables a flexible adoption of the computational budgets.

We summarize our contributions as follows:
\begin{compactitem}[$\bullet$]
    \item Observing the pruning sensitivity of each layer to be different, we propose a novel method, SLS, to determine the layer-wise $N$:$M$ sparsity levels.
    \item From the mixture of the pruned models with different computational costs and accuracy, we propose to find a better trade-off with additional controllability at inference time.
    \item By extensive experiments with super-resolution and deblurring, we empirically validate our pruned models generally achieve state-of-the-art performance.
\end{compactitem}

\section{Related Work}
\label{sec:related_works}

\paragraph{Image Restoration.}
Motion blur or low resolution are common artifacts in images and restoring high-quality from such low-quality inputs has been widely studied in computer vision.
In deep learning literature, many neural network architectures have been proposed to mitigate the artifacts.
In image deblurring, multi-scale architectures~\cite{Nah_2017_CVPR,Tao_2018_CVPR,Gao_2019_CVPR}, stacked networks~\cite{Zhang_2019_CVPR_DMPHN,Suin_2020_CVPR_SAPHN,Zamir_2021_CVPR}, recurrent models~\cite{Zhang_2018_CVPR_svrn} were proposed primarily to achieve better restoration quality.
Due to the difficulty of ill-posed problem, such methods have employed complex architectures with high model capacity, leading to slow execution speed.
Similarly, the advances in deep super-resolution from pioneering SRCNN~\cite{dong2015image} were made by studying various network architectures such as deep networks~\cite{kim2016accurate,lim2017enhanced}, attention mechanisms~\cite{zhang2018image,dai2019second,niu2020single}, and dense connections~\cite{zhang2018residual,wang2018esrgan,haris2018deep}.


\begin{figure*}[t]
    \centering
    \subfloat[Visual illustration for $N$:$M$ sparsity]{
        \includegraphics[width=0.35\textwidth]{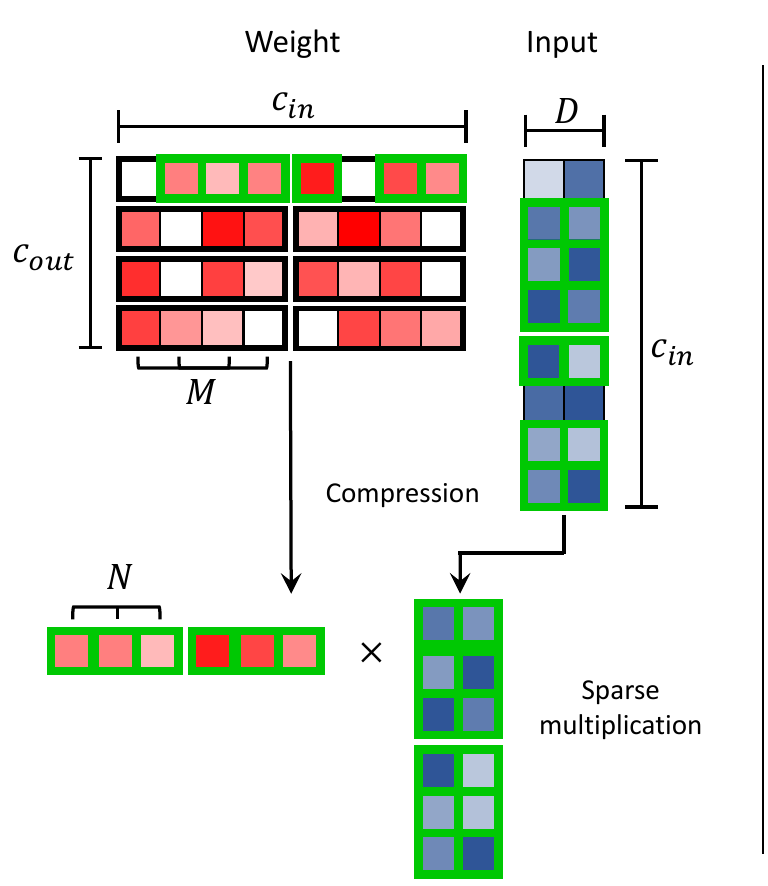}
        \label{fig:n:m}
    }
    \subfloat[Overview of the proposed method]{
        \includegraphics[width=0.65\textwidth]{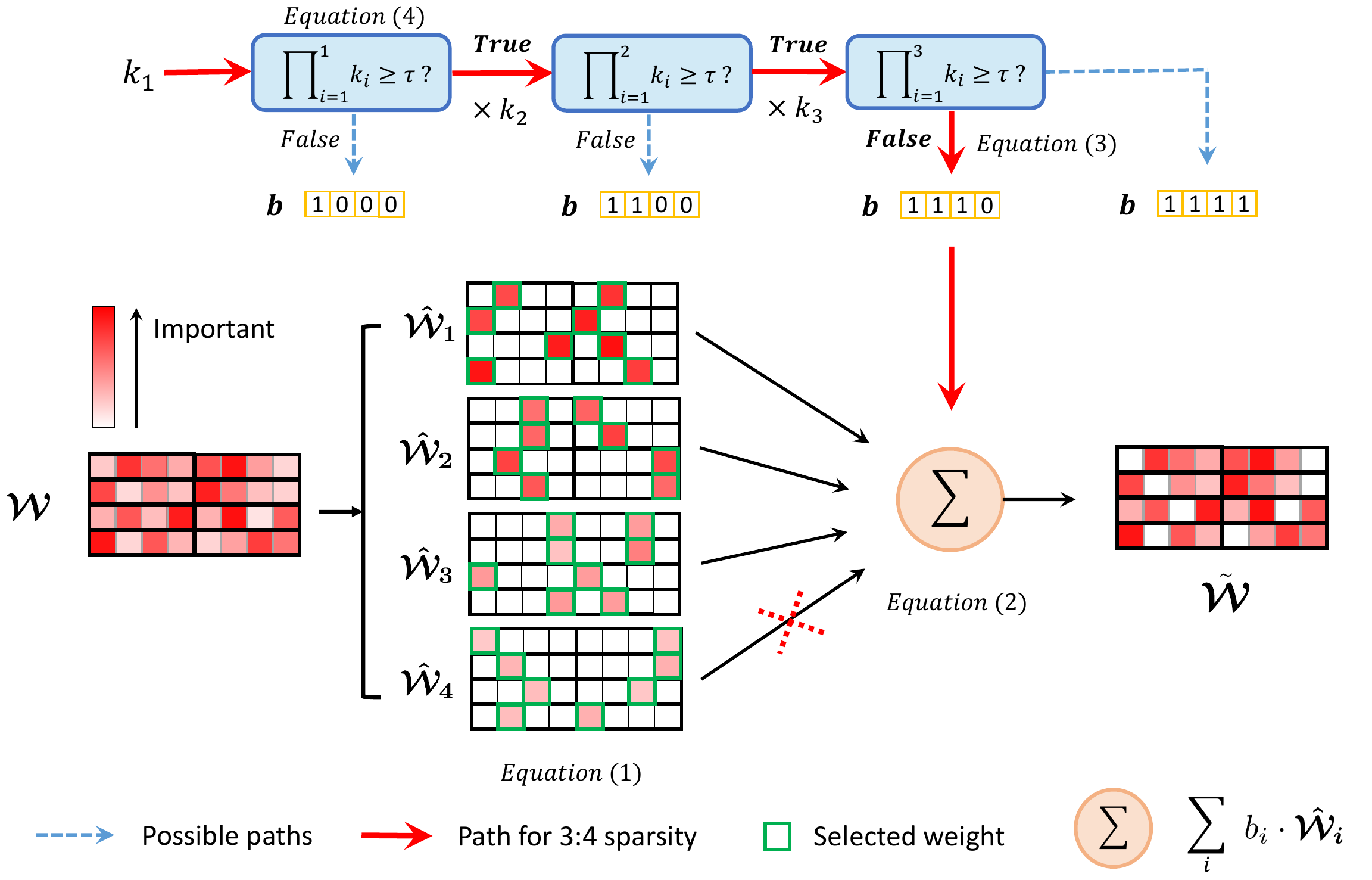}
        \label{fig:proposed_method}
    }
    \figcspace
    \caption{(a) Visual illustration for $N$:$M$ sparsity~\cite{mishra2021accelerating}, where $N=3$ and $M=4$.
    We illustrate the sparse multiplication process only for the first row of the weight, where the non-zero weights and the input features at the corresponding positions are highlighted in green boxes.
    $D$ refers to the spatial dimension of the input feature. (\eg, input height $\times$ input width in a convolutional layer.)
    (b) Overview of the proposed method, SLS, which decomposes the weights into $M$ ($M = 4$ in this case) groups based on the weight magnitude (Equation~\eqref{eq:decompositon}).
    According to Equation~\eqref{eq:pruning_problem}, the final pruned weights are constructed based on the binary mask $\textbf{b}$, each value of which controls whether the corresponding group will be used.
    The binary mask will be generated based on the priority of each importance group, following the operations outlined by Equation~\eqref{eq:ste}~and~\eqref{eq:markov}.
    We illustrate the case when the searched sparsity is 3:4.
    }
    \figspace
\end{figure*}

In order to make deblurring and super-resolution fast, many efforts have been made to design  light-weight architectures.
For deblurring, \cite{Kupyn_2019_ICCV} adopted Inception-ResNet-v2~\cite{szegedy2017inception} and MobileNetV2~\cite{sandler2018mobilenetv2} to build feature pyramid networks.
Also, \cite{Park_2020_ECCV_MTRNN} used a shallow recurrent model and progressively deblurred an image.
Similarly, for super-resolution, \cite{ahn2018fast} and \cite{liu2020residual} proposed to use an effective convolutional module for an efficient yet accurate network.
Furthermore, instead of designing models manually, neural architecture search~(NAS) was adopted~\cite{Chu2019FastAA,Song2019EfficientRD,kim2019fine} to find efficient model structures.
Different from the previous methods designing light-weight architectures, our method makes the existing models efficient by using $N$:$M$ sparsity.

\vspace{-1em}\paragraph{Network Pruning.}
From the early studies in neural networks, the redundant weights that have negligible impact on the final output
have been witnessed and pruning aims to remove such unnecessary components~\cite{lecun1990optimal,hassibi1993optimal}.
In unstructured pruning~\cite{han2015learning,han2016deep,frankle2019lottery,ramanujan2020s}, unimportant individual weight connects were eliminated, with the primary focus on accuracy preservation.
However, due to the irregular sparsity patterns, accelerating the pruned networks requires specially designed hardware, limiting the application in practice~\cite{wen2016learning}.
On the other hand, structured pruning removes groups of the weights (\textit{e.g.,} layers, filters, and channels of a feature map) in the network architecture, leading to easier acceleration on off-the-shelf devices.
Especially, filter or channel pruning methods~\cite{luo2017thinet,he2017channel,lin2020hrank,li2017pruning,liu2017learning,ye2018rethinking,guo2020dmcp,he2019filter,he2020learning,molchanov2019importance,li2020eagleeye} have risen as a popular pruning strategy.
They either train networks to find optimally pruned networks~\cite{liu2017learning,ye2018rethinking,guo2020dmcp,he2020learning,li2020eagleeye} or propose metrics to measure the relative importance of filters or channels in pre-trained models~\cite{luo2017thinet,he2017channel,lin2020hrank,li2017pruning,he2019filter,molchanov2019importance,oh2022batch}.
However, we observed that such coarse-grained structured sparsity can lead to significant damage to the performance in image restoration networks.

Recently, several approaches have been proposed in order to make the structured pruning at a more fine-grained level.
Block-level sparsity with matrix math pipelines~\cite{gray2017gpu} is successfully accelerated with the known pruning structure.
However, it requires to increase the feature size to maintain the original accuracy.
A more fine-grained balanced sparsity is proposed in~\cite{yao2019balanced} which can be accelerated by grouping weights and pruning each group with a uniform sparsity.
Unlike the coarse-grained structured pruning methods, a pruned model by this method can achieve relatively higher accuracy by approximating feature map channels with the remaining weights.
Recently, a similar idea was proposed as 2:4 fine-grained structured sparsity with hardware support on NVIDIA Ampere GPUs~\cite{mishra2021accelerating} and more general $N$:$M$ sparsity configurations were explored in~\cite{zhou2021learning,hubara2021accelerated}.
However, the previous works only consider the uniform $N$:$M$ sparsity levels over all layers, ignoring the layer-wise varying computational costs and contribution to the performance.
In this paper, we further optimize the trade-off between the computational costs and performance by searching for an appropriate $N$:$M$ sparsity level for each layer, specifically effective for extremely pruned networks.


\vspace{-0.2cm}
\section{Proposed Method}
\label{sec:proposed_method}

\subsection{$N$:$M$ sparsity}\label{preliminaries}
A weight tensor with $N$:$M$ sparsity indicates a type of a weight tensor that satisfies the following conditions (See Figure~\ref{fig:n:m}):
(1) The number of input channels is divisible by $M$. (2) Each group of $M$ consecutive weights should have at least $N$ non-zero weights.
With a weight satisfying the above constraints, the weight and input tensor are compressed by ignoring the zero weights and the corresponding input feature values.
Then, the computations of the tensor multiplication between the compressed weight and input tensor are reduced to $\frac{N}{M}$ of the original computations.

\subsection{Overview}
The conventional structured pruning methods~\cite{guo2020dmcp,lin2020hrank,he2017channel,liu2017learning,lin2019towards} remove the predetermined structural units (\eg, filters or channels of a feature map).
The common approach for finding layer-wise pruning ratios in structured pruning methods is to train a score value defined in each pruning unit and remove units with a small score value~\cite{liu2017learning,guo2020dmcp,kang2020operation}.
In the case of $N$:$M$ sparsity pattern, however, it is challenging to determine such structural units because there are many possible configurations for preserving $N$ weights out of $M$ weights.
To overcome the challenge, in Section~\ref{sec:differentiable_search}, we consider the original weight tensor as the sum of $M$ tensors with $1$:$M$ sparsity whose configurations of remaining weights are determined by the magnitude of weight.
We propose to use each sparse tensor as the unit of pruning.
Then, we propose a differentiable sparsity search framework that learns score values for each pruning unit by using a straight-through estimator~\cite{hubara2016binarized}.
We empirically found that the magnitude-based importance and the score value can lead to a conflict with respect to the importance rank of a pruning unit, which brings a substantial performance loss.
In Section~\ref{sec:prioirty}, we eliminate the conflict by ensuring the score value is aligned to the magnitude-based importance.
In Section~\ref{sec:loss}, we present our loss function and propose a loss annealing strategy to control the speed of pruning during training.
In Section~\ref{sec:adaptive_inference}, we propose an adaptive inference method to improve the efficiency-accuracy trade-off.

\vspace{-0.5em}
\subsection{Differentiable $N$:$M$ Sparsity Search}\label{sec:differentiable_search}
Let $\boldsymbol{\mathcal{W}}^l \in \mathbb{R}^{c_{out}^l \times c_{in}^l \times k_h^l \times k_w^l}$ denote the weight tensor in $l$-th convolutional layer.
For notation simplicity, $l$ is omitted unless otherwise noted.
Our goal is to find the effective sparsity level of each layer given the target computational budgets.
To this end, we represent the weight tensor as the sum of sparse tensors with 1:$M$ sparsity:
\begin{equation}\label{eq:decompositon}
        \boldsymbol{\mathcal{W}} = \sum_{i=1}^{M} \hat{\boldsymbol{\mathcal{W}}}_i,
\end{equation}
where $\hat{\boldsymbol{\mathcal{W}}}_i$ is a weight tensor with 1:$M$ sparsity in which only the $i$-th important weight is remained every $M$ consecutive weights.
We define the importance of each weight as its magnitude, which has been widely used in the pruning literature~\cite{zhou2021learning,hubara2021accelerated}.
By doing so, we can view the sparse tensor as a pruning unit in our problem setting.
In other words, $\hat{\boldsymbol{\mathcal{W}}}_i$ is analogous to a filter or a channel in the existing structured pruning methods.
Then, we formulate the pruning ratio search problem as follows:
\begin{equation}\label{eq:pruning_problem}
    \tilde{\boldsymbol{\mathcal{W}}} = \sum_{i=1}^{M} b_i \cdot \hat{\boldsymbol{\mathcal{W}}}_i,
\end{equation}
where $b_i$ is a binary value indicating whether $\hat{\boldsymbol{\mathcal{W}}}_i$ should be removed or not.
To optimize $b_i$ using gradient descent, we adopt straight-through estimator (STE)~\cite{hubara2016binarized}:
\begin{equation}\label{eq:ste}
b_i =
\begin{cases}
S(p_i, \tau) & \mbox {in a forward path}, \\
p_i & \mbox{in a backward path},
\end{cases}
\end{equation}
where $S(\cdot,\tau)$ is a function that returns 1 for the greater value than a threshold $\tau$ and 0 otherwise and $p_i$ is a priority score of $\hat{\boldsymbol{\mathcal{W}}}_i$ that is learned during training.

\subsection{Priority-Ordered Pruning}\label{sec:prioirty}
In section~\ref{sec:differentiable_search}, we introduce the two importance measures for each sparse tensor: the magnitude of weight is for the definition of the pruning unit and the priority score is for learning the pruning ratio.
However, the importance rank indicated by the two measures can be different, by which a pruning unit with weights of larger magnitude can be removed first before one with weights of smaller magnitude.
We found such misalignment leads to substantial performance degradation.
To solve this problem, we propose Priority-Ordered Pruning (POP) method that aligns the two importance measures by design.
Specifically, we define the priority score as follows:
\begin{align}
\label{eq:markov}
       p_{i} &\equiv \prod_{n=1}^{i-1} k_n
\end{align}
where $p_1\equiv1$ and $k_n$ is a trainable parameter that is initialized to 1 and clamped to ensure $0\leq k_n \leq 1$.
By Equation~\eqref{eq:markov}, $p_{i+1} \leq p_{i}$ is guaranteed, so a pruning unit with weights of smaller magnitude is removed first.
Figure~\ref{fig:proposed_method} illustrates the overall process of the proposed learning framework.

\subsection{Loss Function}\label{sec:loss}

For practical usage of the proposed method, we design our pruning framework as a budgeted pruning~\cite{guo2020dmcp,ning2020dsa}, in which a network is pruned to meet the desired target computational budget.
To this end, we first define the computational costs of a convolutional layer with $N$:$M$ sparsity with respect to multiply-accumulate operations (MACs).
Theoretically, the MACs of a convolutional layer with $N$:$M$ sparsity are $\frac{N}{M}$ of the original ones.
Thus, the computational costs of a pruned convolutional layer are defined as follows:
\begin{equation}\label{eq:pruned_comp}
    C_{pruned} = C_{original}\times \frac{\sum_{i=1}^{M} b_i}{M},
\end{equation}
where $C_{original} = (c_{out} \cdot c_{in} \cdot k_h \cdot k_w) \times (H \cdot W)$ denotes the original computational costs of the layer and $H$ and $W$ are the spatial sizes of the output of the layer.
Then, we formulate our loss function as follows:
\begin{equation}\label{eq:loss}
\mathcal{L} = \mathcal{L}_{task} + \lambda_{reg} \sum_{l=1}^L C_{pruned}^l,
\end{equation}
where $L$ is the total number of layers to be pruned, $\mathcal{L}_{task}$ is the task-specific loss function (\eg, L1 loss for image restoration tasks) and $\sum_{l=1}^L C_{pruned}^l$ is the computational regularization loss.
The two loss terms are balanced by the hyper-parameter $\lambda_{reg}$.
Note that the gradient from the computational regularization loss can flow to $k_i$ by using STE in Equation~\eqref{eq:ste}.
Starting from a pretrained network, the network is pruned until satisfying $\sum_{l=1}^L C_{pruned}^l \leq C_{target}$, where $C_{target}$ denotes a target computational budget.
After reaching the target budget, all $k_i$ are frozen and the pruned network is fine-tuned by optimizing the task-specific loss.

We empirically found that when the target computational budget is extremely low (\eg, $C_{target} = 0.1 \times \sum_{l=1}^L C_{original}^l$), $\lambda_{reg}$ should be large enough to reach the target budget.
However, a large $\lambda_{reg}$ can result in an aggressive performance degradation because the network is pruned too fast, which is hard to be recovered even after a fine-tuning process.
To solve this problem, we set $\lambda_{reg}$ as a small value at the beginning of training and gradually increase it according to the pruned rate change.
At every predetermined $K$ epoch, we measure the pruned rate change during the last K epochs and update $\lambda_{reg}$ by following rules:
\begin{equation}\label{eq:loss_annealing}
    \lambda_{reg} =
    \begin{cases}
    \alpha \times \lambda_{reg} & \mbox{if $\Delta \frac{C_{pruned}}{C_{original}} \leq T$},\\
    \lambda_{reg} & \mbox{else},
    \end{cases}
\end{equation}
where $\alpha$ is a hyper-parameter determining how fast $\lambda_{reg}$ is increased and $T$ is the threshold value of pruned rate change for updating $\lambda_{reg}$.
Since the performance of the pruned network is not sensitive to the annealing hyper-parameters ($\alpha$, $T$, and $K$), we fix them for all experiments.

\begin{figure}[t]
\includegraphics[width=1\linewidth]{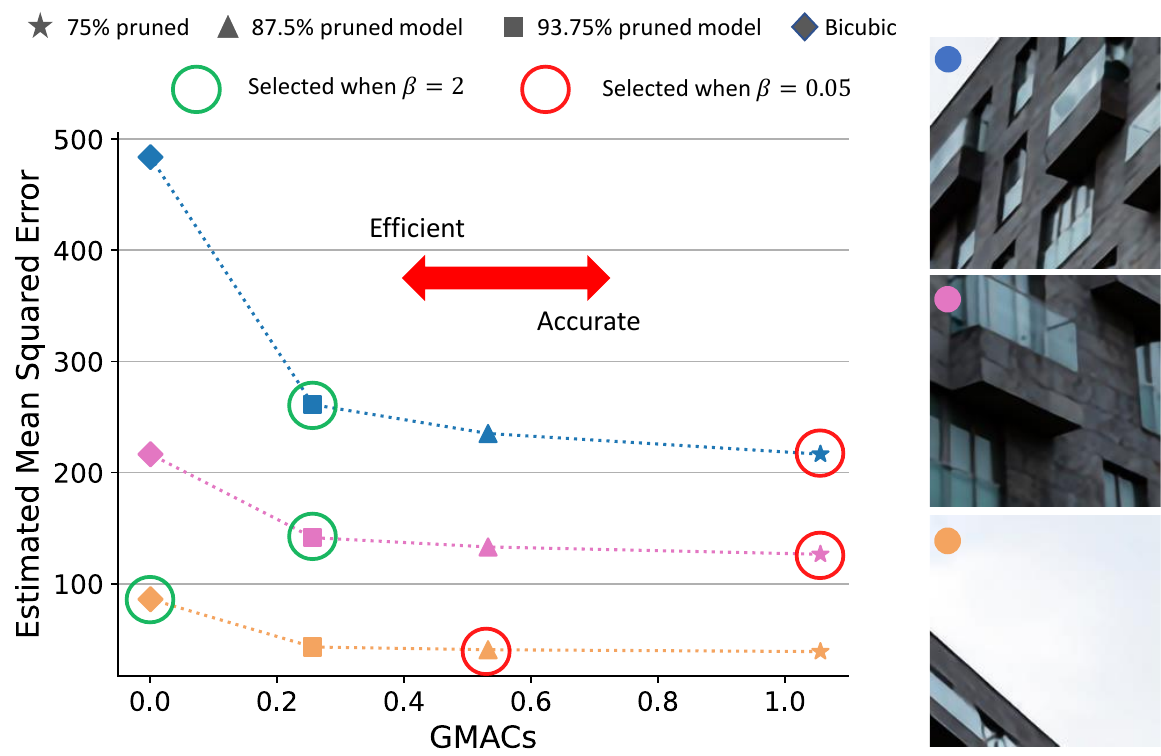}\\
\figcspace
\caption{
Visualization of the proposed adaptive inference method.
Given the 3 images on the right side of the figure and 4 model candidates (75\%, 87.5\%, 93.75\% pruned CARN~\cite{ahn2018fast} and bicubic upsampler), the trained MSE estimators estimate MSE between the restored image and the ground truth.
Then, our method selects one model by Equation~\eqref{eq:adaptive_inference}.
By adjusting $\beta$, one can determine whether to focus on efficiency or accuracy.
}
\label{fig:adaptive_inference}
\figspace
\end{figure}

\subsection{Adaptive Inference}\label{sec:adaptive_inference}
Each region in a low-quality image often has different restoration difficulties.
For example, in the case of image super-resolution tasks, flat areas such as the sky can be easily restored by using only a few computational resources.
Also, in order to restore large images (\eg, 2K or 4K), it can be inevitable to process the whole image by dividing it into small patches due to the resource constraints~\cite{kong2021classsr}.
From this motivation, we propose an adaptive inference method that determines which pruned models, trained by SLS, to use according to the restoration difficulty of an input image patch at inference time.

To quantify the restoration difficulty of a patch, we assume that the more an image patch is hard to be restored, the larger error between the ground truth and the restored result.
Since the ground truth is not available at inference time, we use a light-weight convolutional neural network that can estimate the mean squared error (MSE) between the ground truth and the restored result from a target model.
Given several models trained by SLS with different target computational budgets, we train the MSE estimators for each candidate model using training datasets.

Formally, given an image patch $\boldsymbol{x}$, our adaptive inference method selects a model among the candidates that is obtained by the following operation:
\begin{equation}
\underset{i}{\mathrm{argmax}} \: \frac{C_1-C_i}{C_1} \times \beta + \frac{f_n(\boldsymbol{x}) -f_i(\boldsymbol{x})}{f_n(\boldsymbol{x})},
\label{eq:adaptive_inference}
\end{equation}
where $C_{i}$ and $f_{i}(\cdot)$ indicate the computational costs and the estimated MSE with respect to the $i$-th candidate model and $n$ is the total number of candidates.
We sort the index of each model with respect to the computational costs ($C_{i+1} \leq C_{i}$).
By maximizing the two terms in Equation~\eqref{eq:adaptive_inference}, our method tries to find the most efficient yet accurate model given $\boldsymbol{x}$.
The hyper-parameter $\beta$ enables flexible control of the computational costs of the selected model by making focus on either the computational costs or the performance.
Figure~\ref{fig:adaptive_inference} visualizes the proposed adaptive inference method.

\vspace{-0.2cm}
\section{Experiments}
\label{sec:experiments}
\subsection{Dataset and Models}
To validate the effectiveness of the proposed methods, we conduct experiments on image deblurring and super-resolution tasks.
For image deblurring, we perform pruning on 3 model architectures that vary by the computational cost and the restoration accuracy: residual UNet~\cite{nah2022clean}, SRN~\cite{Tao_2018_CVPR} and DMPHN~\cite{Zhang_2019_CVPR_DMPHN}.
GOPRO dataset~\cite{Nah_2017_CVPR} is employed to train and evaluate the deblurring models.
For image super-resolution, we use 3 popular and efficient architectures: EDSR~\cite{lim2017enhanced}, CARN~\cite{ahn2018fast} and RFDN~\cite{liu2020residual}.
We use DIV2K dataset~\cite{Agustsson_2017_CVPR_Workshops} for training, Set14~\cite{set14}, B100~\cite{martin_2001_ICCV} and Urban100~\cite{Huang_CVPR_2015} benchmark datasets for evaluation.

\vspace{-0.2cm}
\subsection{Implementation Details}
To make a fair comparison between different pruning methods, we train the networks with the same amount of iterations.
The total training epochs are 4000 and 600, respectively for deblurring and super-resolution.
Since the methods except for SR-STE~\cite{zhou2021learning} require a pretraining phase before pruning, we allocate half of the training epochs for pretraining and the rest for the pruning process for those methods.
We set the hyper-parameters as $\tau=0.5$, $\alpha=1.1$, $T=0.1$.
$\lambda_{reg}$ is set to $10^{-12}$ and $10^{-10}$ for image deblurring and super-resolution tasks, respectively.
Also, we set $M=32$ to allow extreme pruning ratios.
For more details, please refer to the supplementary material.

\begin{table}[t]
   \caption{Image deblurring performance comparisons on GOPRO dataset~\cite{Nah_2017_CVPR}.
   } 
   \vspace{-0.1cm}
   \centering
   \scalebox{0.7}{
       \begin{threeparttable}
       \begin{tabular}{clccc}
           \toprule[\heavyrulewidth]
                  Model & Method & GMACs & Num. Param. & PSNR$_\uparrow$ / SSIM$_\uparrow$ / LPIPS$_\downarrow$\\
           \midrule[\heavyrulewidth] 
           \multirow{14}{*}{UNet}& Unpruned & 458.04 & 6.79M & 29.46 / 0.8837 / 0.1686\\
           & One-shot (2:4) & 230.84 & 3.40M & 29.55 / 0.8849 / 0.1662\\
           \cline{2-5}
           & Filter pruning & 115.42 & 1.70M & 28.79 / 0.8692 / 0.1893\\
           & One-shot (8:32) & 117.24 & 1.70M & 29.19 / 0.8771 / 0.1795\\
           & SR-STE (8:32) & 117.24 & 1.70M & 28.85 / 0.8691 / 0.1860\\
           & \cellcolor{orange!20}SLS (Ours) & \cellcolor{orange!20}116.64 & \cellcolor{orange!20}1.55M  & \cellcolor{orange!20}{\textbf{29.37}} / \cellcolor{orange!20}{\textbf{0.8811}} / \cellcolor{orange!20}{\textbf{0.1740}}\\
           \cline{2-5}
           & Filter pruning & 65.27 & 956.74K & 28.31 / 0.8570 / 0.2070\\
           & One-shot (4:32) & 60.44 & 851.38K & 28.64 / 0.8646 / 0.1985\\
           & SR-STE (4:32) & 60.44 & 851.38K & 28.33 / 0.8571 / 0.2070\\
           & \cellcolor{orange!20}SLS (Ours) & \cellcolor{orange!20}60.31 & \cellcolor{orange!20}797.30K  & \cellcolor{orange!20}{\textbf{28.98}} / \cellcolor{orange!20}{\textbf{0.8726}} / \cellcolor{orange!20}{\textbf{0.1870}}\\
           \cline{2-5}
           & Filter pruning & 29.31 & 425.86K & 27.65 / 0.8398 / 0.2325\\
           & One-shot (2:32) & 32.04 & 427.19K & 27.79 / 0.8430 / 0.2345\\
           & SR-STE (2:32) & 32.04 & 427.19K & 27.77 / 0.8431 / 0.2271\\
           & \cellcolor{orange!20}SLS (Ours) & \cellcolor{orange!20}30.91 & \cellcolor{orange!20}397.55K  & \cellcolor{orange!20}{\textbf{28.36}} /  \cellcolor{orange!20}{\textbf{0.8573}} / \cellcolor{orange!20}{\textbf{0.2117}}\\
           \bottomrule[\heavyrulewidth]
           \multirow{14}{*}{SRN}& Unpruned & 1200.51 & 7.09M & 30.28 / 0.9021 / 0.1310\\
           & One-shot (2:4) & 605.35 & 3.55M & 30.53 / 0.9065 / 0.1264\\
           \cline{2-5}
           & Filter pruning & 302.67 & 1.77M & 29.76 / 0.8915 / 0.1467\\
           & One-shot (8:32) & 307.77 & 1.78M & 30.34 / 0.9030 / 0.1313\\
           & SR-STE (8:32) & 307.77 & 1.78M & 29.91 / 0.8942 / 0.1407\\
           & \cellcolor{orange!20}SLS (Ours) & \cellcolor{orange!20}306.84 & \cellcolor{orange!20}1.72M  & \cellcolor{orange!20}\textbf{30.45} / \cellcolor{orange!20}\textbf{0.9051} / \cellcolor{orange!20}\textbf{0.1283}\\
           \cline{2-5}
           & Filter pruning & 171.21 & 998.98K & 29.33 / 0.8821 / 0.1603\\
           & One-shot (4:32) & 158.98 & 895.00K & 29.87 / 0.8944 / 0.1427\\
           & SR-STE (4:32) & 158.98 & 895.00K & 29.35 / 0.8839 / 0.1545\\
           & \cellcolor{orange!15}SLS (Ours) & \cellcolor{orange!20}157.41 & \cellcolor{orange!20}897.97K  & \cellcolor{orange!20}\textbf{30.21} / \cellcolor{orange!20}\textbf{0.9006} / \cellcolor{orange!20}\textbf{0.1351}\\
           \cline{2-5}
           & Filter pruning & 76.94 & 444.87K & 28.70 / 0.8691 / 0.1811\\
           & One-shot (2:32) & 84.59 & 252.50K & 29.04 / 0.8766 / 0.1688\\
           & SR-STE (2:32) & 84.59 & 252.50K & 28.83 / 0.8723 / 0.1747\\
           & \cellcolor{orange!20}SLS (Ours) & \cellcolor{orange!20}84.52 & \cellcolor{orange!20}489.81K  & \cellcolor{orange!20}\textbf{29.75} / \cellcolor{orange!20}\textbf{0.8918} / \cellcolor{orange!20}\textbf{0.1470}\\
           \bottomrule[\heavyrulewidth]
           \multirow{14}{*}{DMPHN}& Unpruned & 994.48 & 8.05M & 31.22 / 0.9164 / 0.1243\\
           & One-shot (2:4) & 501.86 & 4.03M & 31.43 / 0.9196 / 0.1192\\
           \cline{2-5}
           & Filter pruning & 250.94 & 2.02M & 30.47 / 0.9043 / 0.1417\\
           & One-shot (8:32) & 255.55 & 2.02M & 31.05 / 0.9137 / 0.1292\\
           & SR-STE (8:32) & 255.55 & 2.02M & 30.63 / 0.9058 / 0.1406\\
           & \cellcolor{orange!20}SLS (Ours) & \cellcolor{orange!20}254.64 & \cellcolor{orange!20}2.24M  & \cellcolor{orange!20}{\textbf{31.26}} / \cellcolor{orange!20}{\textbf{0.9170}} / \cellcolor{orange!20}{\textbf{0.1242}}\\
           \cline{2-5}
           & Filter pruning & 142.02 & 1.14M & 29.58 / 0.8872 / 0.1639\\
           & One-shot (4:32) & 132.40 & 1.01M & 30.13 / 0.8984 / 0.1504\\
           & SR-STE (4:32) & 132.40 & 1.01M  & 29.79 / 0.8916 / 0.1559\\
           & \cellcolor{orange!20}SLS (Ours) & \cellcolor{orange!20}132.26 & \cellcolor{orange!20}1.20M  & \cellcolor{orange!20}{\textbf{30.79}} / \cellcolor{orange!20}{\textbf{0.9097}} / \cellcolor{orange!20}{\textbf{0.1343}}\\
           \cline{2-5}
           & Filter pruning & 63.90 & 506.41K & 28.61 / 0.8662 / 0.1908\\
           & One-shot (2:32) & 70.82 & 506.88K & 27.60 / 0.8393 / 0.2348\\
           & SR-STE (2:32) & 70.82 & 506.88K & 28.60 / 0.8678 / 0.1831\\
           & \cellcolor{orange!20}SLS (Ours) & \cellcolor{orange!20}70.29 & \cellcolor{orange!20}603.47K  & \cellcolor{orange!20}{\textbf{30.04}} /  \cellcolor{orange!20}{\textbf{0.8967}} / \cellcolor{orange!20}{\textbf{0.1525}}\\
           \bottomrule[\heavyrulewidth]
       \end{tabular}
       \end{threeparttable}
   }
   \label{tab:deblurring_main_results}
   \tabspace
\end{table}

\begin{table}[t]
   \caption{Image super-resolution performace (PSNR$_\uparrow$) comparisons on benchmark datasets with the scaling factor of 4.
   } 
   \vspace{-0.1cm}
   \centering
   \scalebox{0.7}{
       \begin{threeparttable}
       \begin{tabular}{clccc}
           \toprule[\heavyrulewidth]
                  Model & Method & GMACs & Num. Param.
                 & Set14 / B100 / Urban100\\
           \midrule[\heavyrulewidth] 
           \multirow{14}{*}{EDSR}& Unpruned & 114.49 & 1.52M & 28.58 / 27.56 / 26.04\\
           & One-shot (2:4) & 58.22 & 765.00K & 28.56 / 27.55 / 26.01\\
           \cline{2-5}
           & Filter pruning & 29.11 & 380.93K & 28.44 / 27.48 / 25.75\\
           & One-shot (8:32) & 30.09 & 345.50K & \textbf{28.49} / 27.50 / 25.83\\
           & SR-STE (8:32) & 30.09 & 345.50K & 28.44 / 27.46 / 25.73\\
           & \cellcolor{orange!20}SLS (Ours) & \cellcolor{orange!20}29.56 & \cellcolor{orange!20}363.39K  & \cellcolor{orange!20}\textbf{28.49} / \cellcolor{orange!20}\textbf{27.51} / \cellcolor{orange!20}\textbf{25.84}\\
           \cline{2-5}
           & Filter pruning & 16.56 & 214.85K & 28.35 / 27.41 / 25.59\\
           & One-shot (4:32) & 16.02 & 174.75K & 28.34 / 27.41 / 25.52\\
           & SR-STE (4:32) & 16.02 & 174.75K & 28.33 / 27.39 / 25.51\\
           & \cellcolor{orange!20}SLS (Ours) & \cellcolor{orange!20}15.62 & \cellcolor{orange!20}190.59K  & \cellcolor{orange!20}\textbf{28.38} / \cellcolor{orange!20}\textbf{27.43} / \cellcolor{orange!20}\textbf{25.63}\\
           \cline{2-5}
           & Filter pruning & 7.52 & 96.00K & 28.13 / 27.20 / 25.17\\
           & One-shot (2:32) & 8.98 & 89.38K & 28.10 / 27.23 / 25.11\\
           & SR-STE (2:32) & 8.98 & 89.38K & 28.13 / 27.27 / 25.22\\
           & \cellcolor{orange!20}SLS (Ours) & \cellcolor{orange!20}8.65 & \cellcolor{orange!20}97.28K  & \cellcolor{orange!20}\textbf{28.22} / \cellcolor{orange!20}\textbf{27.32} / \cellcolor{orange!20}\textbf{25.31}\\
           \bottomrule[\heavyrulewidth]
           \multirow{14}{*}{CARN}& Unpruned & 91.22 & 1.11M & 28.49 / 27.49 / 25.82\\
           & One-shot (2:4) & 46.63 & 565.00K & 28.49 / 27.51 / 25.86\\
           \cline{2-5}
           & Filter pruning & 23.31 & 279.46K & 28.37 / 27.43 / 25.64\\
           & One-shot (8:32) & 24.20 & 278.76K & 28.44 / 27.47 / 25.74\\
           & SR-STE (8:32) & 24.20 & 278.76K & 28.40 / 27.42 / 25.65\\
           & \cellcolor{orange!20}SLS (Ours) & \cellcolor{orange!20}24.09 & \cellcolor{orange!20}276.34K  & \cellcolor{orange!20}\textbf{28.46} / \cellcolor{orange!20}\textbf{27.48} / \cellcolor{orange!20}\textbf{25.79}\\
           \cline{2-5}
           & Filter pruning & 13.31 & 157.75K & 28.26 / 27.32 / 25.36\\
           & One-shot (4:32) & 13.02 & 140.24K & 28.24 / 27.32 / 25.36\\
           & SR-STE (4:32) & 13.02 & 140.24K & 28.23 / 27.33 / 25.43\\
           & \cellcolor{orange!20}SLS (Ours) & \cellcolor{orange!20}13.02 & \cellcolor{orange!20}140.27K  & \cellcolor{orange!20}\textbf{28.39} / \cellcolor{orange!20}\textbf{27.41} / \cellcolor{orange!20}\textbf{25.61}\\
           \cline{2-5}
           & Filter pruning & 6.08 & 70.61K & 27.98 / 27.15 / 24.93\\
           & One-shot (2:32) & 7.44 & 70.97K & 27.86 / 27.08 / 24.78\\
           & SR-STE (2:32) & 7.44 & 70.97K & 28.03 / 27.19 / 25.09\\
           & \cellcolor{orange!20}SLS (Ours) & \cellcolor{orange!20}7.26 & \cellcolor{orange!20}67.18K  & \cellcolor{orange!20}\textbf{28.22} / \cellcolor{orange!20}\textbf{27.32} / \cellcolor{orange!20}\textbf{25.35}\\
           \bottomrule[\heavyrulewidth]
           \multirow{14}{*}{RFDN}& Unpruned & 39.86 & 828.75K & 28.52 / 27.51 / 25.91\\
           & One-shot (2:4) & 20.02 & 416.25K & 28.54 / 27.51 / 25.92\\
           \cline{2-5}
           & Filter pruning & 10.44 & 214.77K & 16.50 / 17.32 / 15.52\\
           & One-shot (8:32) & 10.10 & 210.00K & 28.46 / 27.47 / 25.73\\
           & SR-STE (8:32) & 10.10 & 210.00K & 28.33 / 27.41 / 25.59\\
           & \cellcolor{orange!20}SLS (Ours) & \cellcolor{orange!20}10.05 & \cellcolor{orange!20}240.17K  & \cellcolor{orange!20}\textbf{28.50} / \cellcolor{orange!20}\textbf{27.50} / \cellcolor{orange!20}\textbf{25.84}\\
           \cline{2-5}
           & Filter pruning & 6.05 & 123.84K & 15.76 / 16.61 / 14.78\\
           & One-shot (4:32) & 5.17 & 106.88K & 27.99 / 27.16 / 24.93\\
           & SR-STE (4:32) & 5.17 & 106.88K & 28.19 / 27.31 / 25.30\\
           & \cellcolor{orange!20}SLS (Ours) & \cellcolor{orange!20}5.16 & \cellcolor{orange!20}139.60K  & \cellcolor{orange!20}\textbf{28.41} / \cellcolor{orange!20}\textbf{27.43} / \cellcolor{orange!20}\textbf{25.63}\\
           \cline{2-5}
           & Filter pruning & 2.85 & 57.74K & 15.01 / 15.82 / 14.03\\
           & One-shot (2:32) & 2.66 & 55.31K & 27.49 / 27.06 / 24.72\\
           & SR-STE (2:32) & 2.66 & 55.31K & 28.02 / 27.21 / 25.08\\
           & \cellcolor{orange!20}SLS (Ours) & \cellcolor{orange!20}2.66 & \cellcolor{orange!20}77.49K  & \cellcolor{orange!20}\textbf{28.22} / \cellcolor{orange!20}\textbf{27.32} / \cellcolor{orange!20}\textbf{25.31}\\
           \bottomrule[\heavyrulewidth]
       \end{tabular}
       \end{threeparttable}
   }
   \label{tab:sr_main_results}
    \tabspace
\end{table}

\subsection{Quantitative Comparison}
In Table~\ref{tab:deblurring_main_results} and \ref{tab:sr_main_results}, we present the computational costs (GMACs) and the image restoration performance of the pruned models on image deblurring and super-resolution tasks, respectively.
For each model, we train them using different computational budgets, 1/4, 1/8, and 1/16 of the original costs.
Our method, SLS, is compared with the existing pruning methods, filter pruning~\cite{li2017pruning}, one-shot $N$:$M$ pruning~\cite{mishra2021accelerating} and SR-STE~\cite{zhou2021learning}, with respect to PSNR, SSIM, and LPIPS.
Under almost the same GMACs, SLS consistently achieves the best image restoration performance across all tasks and model architectures.
Especially, SLS outperforms the other methods by a large margin at extremely pruned cases.
These results show that our method can achieve a better trade-off between the computational costs and restoration performance by searching for the effective layer-wise $N$:$M$ sparsity levels.


\begin{figure*}[ht]
    \begin{center}\centering
    \setlength{\tabcolsep}{0.05cm}
    \hspace*{-\tabcolsep}
    \begin{subfigure}[b]{1\textwidth}
    \centering
    \begin{tabular}{ccccc}
    &&\footnotesize Super-Resolution&&\vspace{0.1cm}\\
    \includegraphics[width=0.18\linewidth]{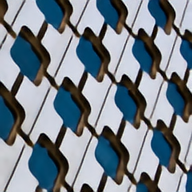} & 
    \includegraphics[width=0.18\linewidth]{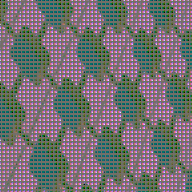} &
    \includegraphics[width=0.18\linewidth]{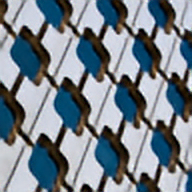} & 
    \includegraphics[width=0.18\linewidth]{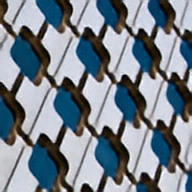} &
    \includegraphics[width=0.18\linewidth]{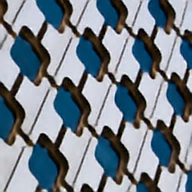}\\
    
    \includegraphics[width=0.18\linewidth]{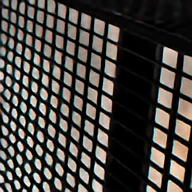} & 
    \includegraphics[width=0.18\linewidth]{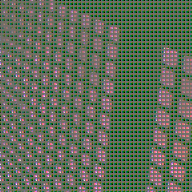} &
    \includegraphics[width=0.18\linewidth]{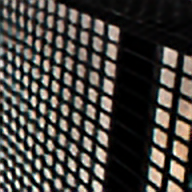} & 
    \includegraphics[width=0.18\linewidth]{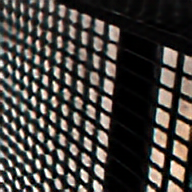} &
    \includegraphics[width=0.18\linewidth]{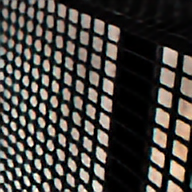} 
    \\
    \footnotesize Unpruned & \footnotesize Filter Pruning &\footnotesize One-shot (2:32) &\footnotesize SR-STE (2:32) &\footnotesize SLS (Ours) \\
    \end{tabular}
    \vspace{0.05cm}
    \label{figure:sr_qualitative}
    \end{subfigure}
    \begin{subfigure}[b]{1\textwidth}
    \centering
    \vspace{+0.0cm}
    \begin{tabular}{ccccc}
    &&\footnotesize Deblurring&&\vspace{0.1cm}\\
    \includegraphics[width=0.18\linewidth]{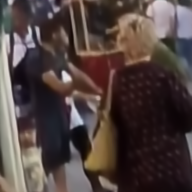} & 
    \includegraphics[width=0.18\linewidth]{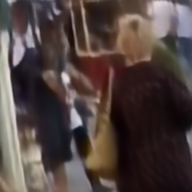} &
    \includegraphics[width=0.18\linewidth]{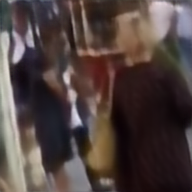} & 
    \includegraphics[width=0.18\linewidth]{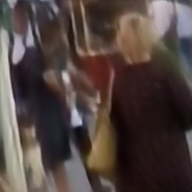} &
    \includegraphics[width=0.18\linewidth]{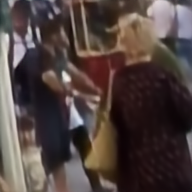}\\
    \includegraphics[width=0.18\linewidth]{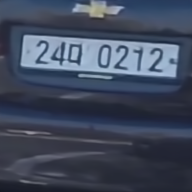} & 
    \includegraphics[width=0.18\linewidth]{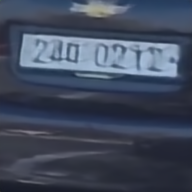} &
    \includegraphics[width=0.18\linewidth]{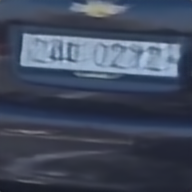} & 
    \includegraphics[width=0.18\linewidth]{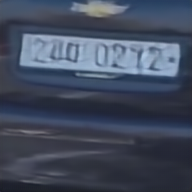} &
    \includegraphics[width=0.18\linewidth]{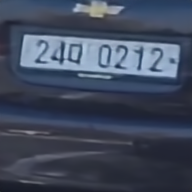}  \vspace{-0.1cm}
    \\
    \footnotesize Unpruned & \footnotesize Filter Pruning &\footnotesize One-shot (2:32) &\footnotesize SR-STE (2:32) &\footnotesize SLS (Ours) \vspace{-0.1cm} \\
    \end{tabular}
    \label{figure:deblur_qualitative}
    \end{subfigure}
    \figcspace
    \captionof{figure}{Qualitative comparisons with the existing pruning methods. The first two rows show image super-resolution results from RFDN with the scaling factor of 4. The last two rows show image deblurring results from DMPHN. For each task, all pruned models have almost the same computational costs (1/16 of the original value) with respect to GMACs.
    }
    \vspace{-1em}
    \figspace
    \label{fig:qualitative}
    \end{center}
\end{figure*}

\subsection{Qualitative Comparison}
We present the qualitative results in Figure~\ref{fig:qualitative}.
In the case of image super-resolution tasks (the first two rows), we observe that the models trained by SLS can restore more sharp and clear textures, compared to the results from the other methods.
Notably, the results from the filter pruning suffer checkerboard artifacts since pruning the last pixel shuffle upscaling layer~\cite{Shi_2016_CVPR} results in sparse pixel values.
Similarly, the results on image deblurring tasks (the last two rows) show that the models trained by SLS can restore the detailed textures and cleaner car plates with better readability while other pruning methods fail to reconstruct such high-frequency details.
The overall results show that under the same computational budgets, models pruned by SLS achieve perceptually satisfying performance.

\begin{table}[h!]
\renewcommand{\cmark}{\textcolor{MyGreen}{\ding{51}}}
   \caption{Ablation studies on the proposed methods. 
   In the 3rd column, the two numbers indicate the update periods (iterations) for updating $\hat{\boldsymbol{\mathcal{W}}}^l_i$ in RFDN and UNet, respectively.
   For RFDN, we use Urban100 dataset.
   We set the target budget as $\frac{1}{8}$ of the original computational costs.
   } 
   \tabcspace
   \centering
   \scalebox{0.8}{
       \begin{threeparttable}
       \begin{tabular}{ccccc}
           \toprule[\heavyrulewidth]
                  \multirow{2}{*}{POP} & \multirow{2}{*}{$\lambda_{reg}$ annealing} & \multirow{1}{*}{Update period} & \multicolumn{2}{c}{PSNR$_\uparrow$}\\
                  \cmidrule{4-5}
                  &&(Iterations)&RFDN & UNet\\
           \midrule[\heavyrulewidth]
            \xmark & \cmark & 1000/131 & 25.46 & 28.57 \\ 
            \cmidrule{1-5}
            \cmark & \xmark & 1000/131 & 25.59 & 28.84\\
            \cmidrule{1-5}
            \cmark & \cmark & No update & 25.46 & 28.74 \\
            \cmark & \cmark & 1/1  & 25.61 & 28.88 \\ 
            \cmark & \cmark & 10000/1310 & 25.61 & 28.92\\
            \cmidrule{1-5}
            \cellcolor{orange!20}\cmark & \cellcolor{orange!20}\cmark & \cellcolor{orange!20}1000/131 & \cellcolor{orange!20}\textbf{25.63} & \cellcolor{orange!20}\textbf{28.98}\\
           \bottomrule[\heavyrulewidth]
       \end{tabular}
       \end{threeparttable}}
   \label{tab:abldation}
   \vspace{0.5em}
    \tabspace
\end{table}

\begin{figure*}[t]
\begin{center}
\subfloat[Visualization of the searched layer-wise computational costs (MACs)]{
\includegraphics[width=0.48\textwidth]{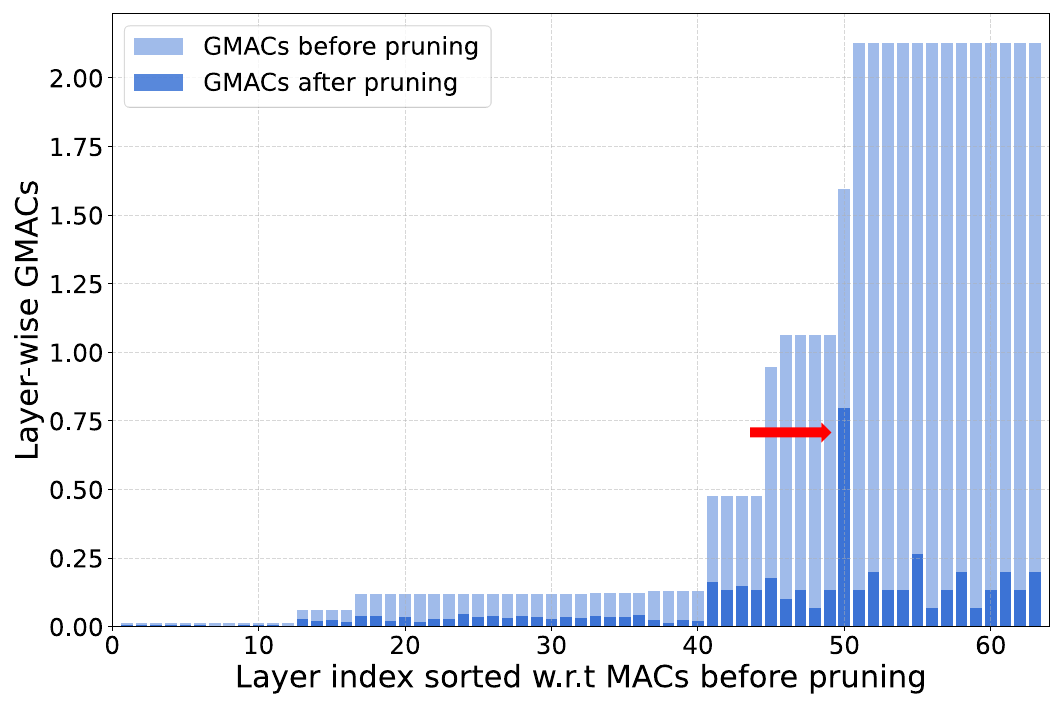}
\label{fig:searched_pruned_rate}
}
\subfloat[Visualization of the searched layer-wise $N$ (when $M=32$)]{
\includegraphics[width=0.48\textwidth]{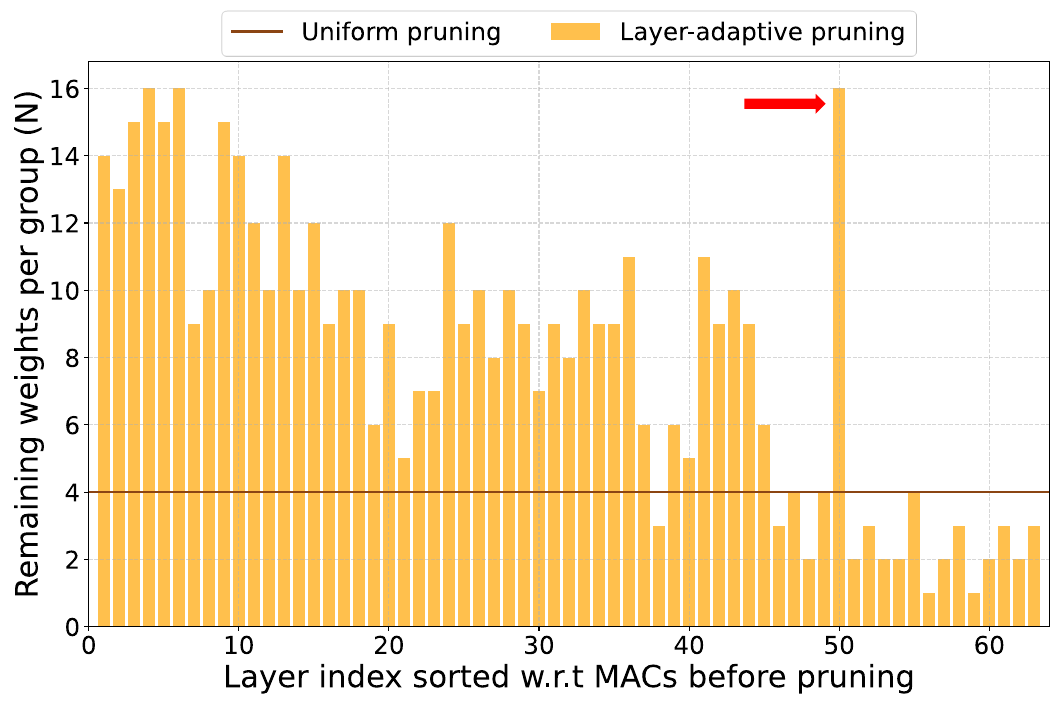}
\label{fig:searched_n:m}
}
\end{center}
\figcspace
\figcspace
\vspace{-0.2cm}
\caption{Analysis on the searched layer-wise $N$:$M$ sparsity.
The results are obtained by training RFDN model with $C_{target}=\frac{1}{8}C_{original}$.
We visualize the layer-wise pruning ratios in terms of (a) GMACs and (b) $N$.
The layer is sorted with respect to MACs before pruning.
The highlighted bar by the red arrow indicates the result of the last upsampling layer in RFDN.
The brown line in (b) is for the comparison with the uniform pruning~\cite{mishra2021accelerating,zhou2021learning}.
}
\figspace
\vspace{-0.2cm}
\label{fig:layer_wise_n:m_visualization}
\end{figure*}

\subsection{Finding Optimal Pruning Ratio for Each Layer}
Different from the previous methods that set a uniform $N$:$M$ sparsity in all layers, our SLS finds the sparsity level for each layer via learning.
In Figure~\ref{fig:layer_wise_n:m_visualization}, we visualize the searched level of sparsity of each layer.
The results show that the searched $N$ have a tendency to decrease when the computational costs of the corresponding layer become heavy.
Interestingly, we found that this tendency is not shown in the last upscaling layer in RFDN (indicated as the 50-th layer in the figures).
We empirically found that the last upscaling layer has a significant impact on the performance since it is directly related to the final output.
Thus, considering the better restoration performance in Table~\ref{tab:sr_main_results}, these results demonstrate that SLS finds more effective pruning ratios for each layer by taking into account both the computational costs and the contribution to the performance of each layer.

\begin{figure}[t]
\centering
\includegraphics[width=1\linewidth]{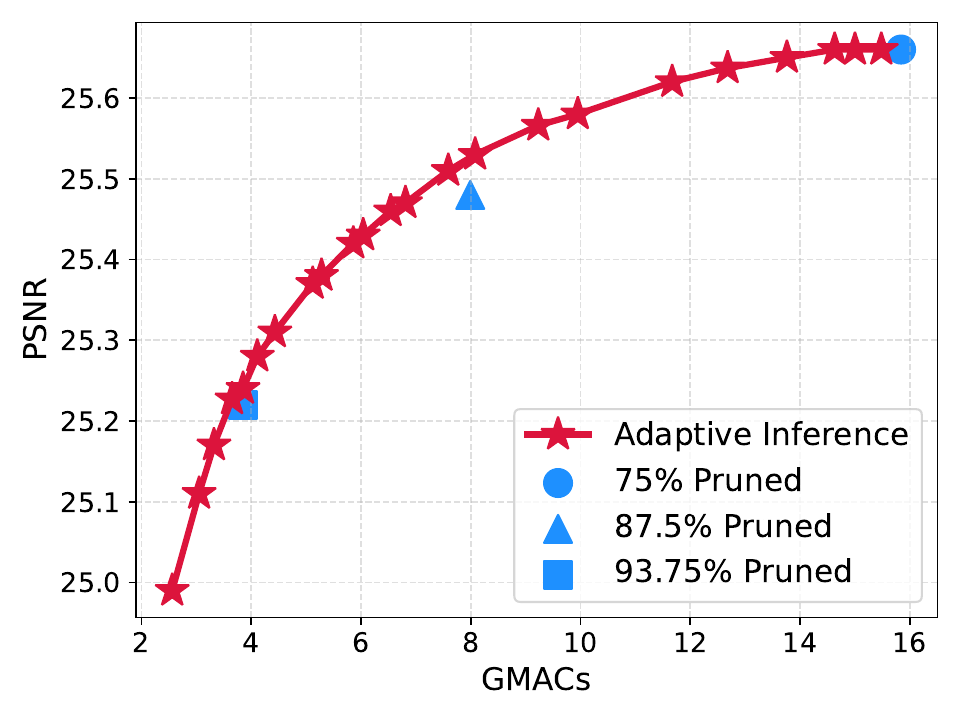}
\figcspace
\vspace{-0.5cm}
\caption{Results of the proposed adaptive inference method on Urban100 dataset with the scale factor of 4.
We use the bicubic upsampler and three CARN models trained by SLS with different target budgets.
}
\label{fig:adaptive_inference_results}
\vspace{-0.3cm}
\figspace
\end{figure}

\subsection{Ablation Studies}
To validate the effectiveness of each component in SLS, we conduct ablation studies and the results are shown in Table~\ref{tab:abldation}.
To investigate the effect of Priority-Ordered Pruning (POP), we defined the priority score values $p$ not as the cumulative product of auxiliary trainable parameters as in Equation~\eqref{eq:markov} but as independent trainable parameters.
The results show that there is a large performance loss when POP is not used, indicating that aligning the two importance measures (the magnitude of weights and the priority scores) is essential.
For the ablation study on $\lambda_{reg}$ annealing, we set $\lambda_{reg}$ as a large value that is finally found when the annealing strategy is used and train the models with it.
The results demonstrate that $\lambda_{reg}$ annealing brings an additional performance gain by controlling the speed of pruning process according to the current pruned rate.
In Equation~\eqref{eq:decompositon}, we group the weight tensor into several sparse tensors by using the magnitude of weights.
Since the weights change during training, we should update the sparse tensors, $\hat{\boldsymbol{\mathcal{W}}}_i$.
To find an appropriate update period, we train models with different update periods.
As expected, the pruned models suffer significant performance degradation when there is no update.
Also, updating $\hat{\boldsymbol{\mathcal{W}}}_i$ at every iteration is not helpful, so we set the update period as 10000 and 1310 iterations (10 epochs) for super-resolution and deblurring, respectively.

\vspace{-0.1cm}
\subsection{Adaptive Inference}\label{sec:adaptive_inference}
Figure~\ref{fig:adaptive_inference_results} shows the results of the proposed adaptive inference method on image super-resolution tasks.
An input image is divided into several patches and each patch is restored by the selected model by Equation~\eqref{eq:adaptive_inference}.
As shown in the figure, our adaptive inference scheme not only enables the detailed control of computational budgets but also improves the trade-off between the computational costs and the restoration performance in terms of PSNR.
For the experimental details and the results on deblurring tasks, please refer to the supplementary material.

\vspace{-0.1cm}
\section{Conclusion}
\label{sec:conclusion}
In this paper, we propose a novel layer-wise pruning ratio search framework, SLS, tailored for $N$:$M$ sparsity.
Our differentiable learning framework is trained end-to-end with the task-specific and the computational regularization loss to determine a more effective degree of sparsity for each layer.
Compared with the previous methods with uniform $N$:$M$ sparsity at all layers, our results achieve state-of-the-art image restoration performance at similar computational budgets.
Furthermore, our adaptive inference scheme facilitates the detailed control of the computational budgets with improved restoration performance.

{\small \paragraph{Acknowledgment.}
This work was supported in part by IITP grant funded by the Korea government (MSIT) [No. 2021-0-01343, Artificial Intelligence Graduate School Program (Seoul National University), and No. 2021-0-02068, Artificial Intelligence Innovation Hub], and in part by AIRS Company in Hyundai  Motor Company \& Kia Motors Corporation through HMC/KIA-SNU AI Consortium.}

\newpage
\newtoggle{isarxiv}
\toggletrue{isarxiv}
\renewcommand{\thetable}{S\arabic{table}}
\renewcommand{\thefigure}{S\arabic{figure}}
\renewcommand{\theequation}{S\arabic{equation}}
\renewcommand{\thesection}{S\arabic{section}}
\setcounter{table}{0}
\setcounter{figure}{0}
\setcounter{equation}{0}
\setcounter{section}{0}

\section*{Appendix}
In this appendix, we elaborate on the validity of our proposed pruning method in detail.
In Seciton~\ref{supp_sec:supp_reg_schedule}, we validate the necessity and efficacy of the proposed regularization loss annealing during training.
In Section~\ref{supp_sec:supp_generalization_to_M}, we show that our search framework is also effective at various $N$:$M$ configurations.
In Section~\ref{supp_sec:supp_adaptive_inference}, we provide additional experimental results on the proposed adaptive inference scheme on the image deblurring task.
Finally, in Section~\ref{supp_sec:details}, we provide the implementation details for reproducibility.

\section{Effectiveness of Scheduling $\lambda_{reg}$}
\label{supp_sec:supp_reg_schedule}
Our pruning framework prunes a pretrained network by jointly optimizing the task loss and the computational amount regularization loss until the pruned model meets the target computational budget.
As discussed in Section~\ref{sec:loss}, a large constant weight for $\lambda_{reg}$ leads to immature pruning.
The induced performance drop is significant even after the following fine-tuning process.
On the other hand, if $\lambda_{reg}$ is smaller, the model fails to meet the target budget.
To overcome this dilemma, we adaptively change $\lambda_{reg}$ by monitoring the progress of the pruning rates in the last few epochs (Eq.~\eqref{eq:loss_annealing}).

\begin{figure}[!t]
    \newcommand{\wwp}{0.95\linewidth}
    \centering
    \includegraphics[width=\wwp]{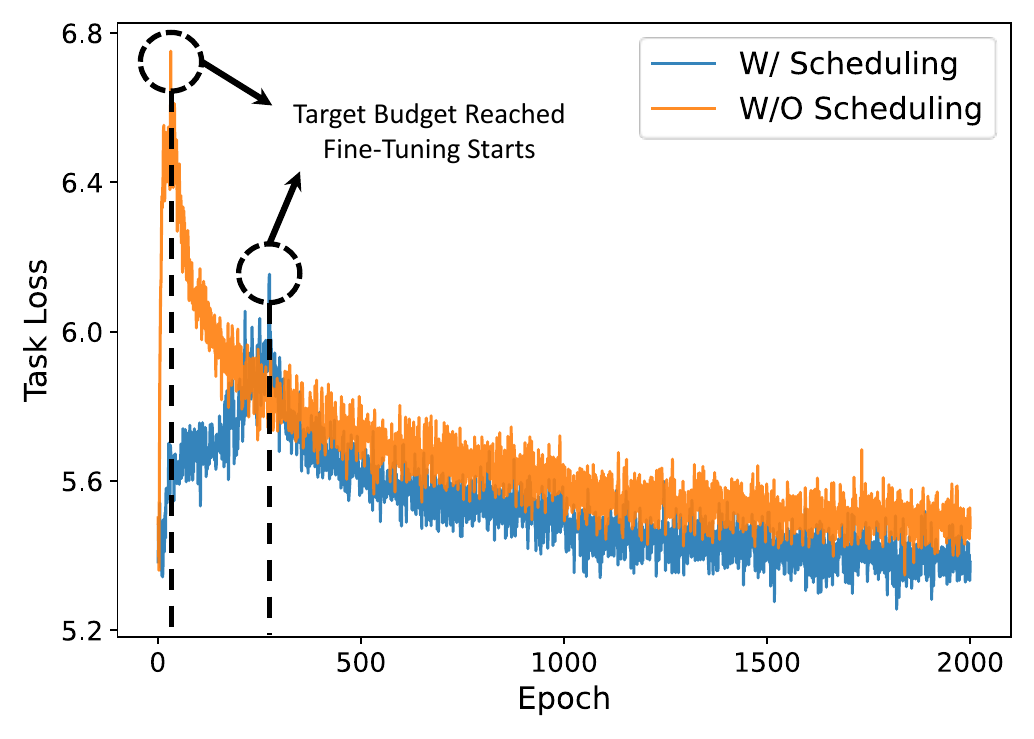}
    \figcspace
    \caption{Task loss curves with (blue curve) and without (orange curve) $\lambda_{reg}$ scheduling. 
    The highlighted parts indicate the moment when the target budget ($\frac{1}{8}$ of the original computational costs) is reached in each case.
    In the case of `W/O Scheduling', we set $\lambda_{reg}$ as the final value of $\lambda_{reg}$ in `W/ Scheduling'.}
    \label{fig:supp_schedule}
\end{figure}

In addition to the quantitative validation in Table~\ref{tab:abldation} in the main manuscript, we contrast the effect of proper $\lambda_{reg}$ scheduling by loss curve as the learning progresses in Figure~\ref{fig:supp_schedule}.
From a constant $\lambda_{reg}$ without the scheduling, the model rapidly deteriorates with the spiking task loss.
In contrast, our scheduling on $\lambda_{reg}$ prunes the model in a moderate speed so that the optimization becomes easier even with a relatively less number of fine-tuning epochs.
These results imply the effectiveness of our $\lambda_{reg}$ scheduling.

\section{Generalization to Various $M$ Values}
\label{supp_sec:supp_generalization_to_M}
In Tables \ref{tab:deblurring_main_results} and \ref{tab:sr_main_results} of the main manuscript, we presented the $N$:$M$ pruning results when $M = 32$ to elaborate the effectiveness of our SLS framework in an extreme pruning ratio up to roughly 93.75\%.
We show that our SLS generalizes to other configurations by quantitative comparisons with various $M$ values.

Tables~\ref{tab:deblurring_supp} and \ref{tab:sr_supp} show the pruning results on image deblurring and super-resolution models for $M \in \{8, 16, 32\}$, respectively.
For all methods, the restoration performance tend to degrade when $M$ is smaller.
Nevertheless, SLS consistently outperforms others on both tasks and for all $M$.
Empirically, SLS is generalized well on various $M$.

\begin{table}[t]
   \caption{Image deblurring performance comparisons on GOPRO dataset with various $M$.
   } 
   \centering
   \resizebox{\linewidth}{!}{
       \begin{threeparttable}
       \begin{tabular}{clccc}
           \toprule[\heavyrulewidth]
                  Model & Method & GMACs & Num. Param. & PSNR$_\uparrow$ / SSIM$_\uparrow$ / LPIPS$_\downarrow$\\
           \midrule[\heavyrulewidth] 
           \multirow{11}{*}{UNet}& Unpruned & 458.04 & 6.79M & 29.46 / 0.8837 / 0.1686\\
           \cline{2-5}
           & Filter pruning & 115.42 & 1.70M & 28.79 / 0.8692 / 0.1893\\
           & One-shot (8:32) & 117.24 & 1.70M & 29.19 / 0.8771 / 0.1795\\
           & SR-STE (8:32) & 117.24 & 1.70M & 28.85 / 0.8691 / 0.1860\\
           & \cellcolor{orange!20}SLS (Ours) & \cellcolor{orange!20}116.64 & \cellcolor{orange!20}1.55M  & \cellcolor{orange!20}{\textbf{29.37}} / \cellcolor{orange!20}{\textbf{0.8811}} / \cellcolor{orange!20}{\textbf{0.1740}}\\
           \cline{2-5}
           & One-shot (4:16) & 117.24 & 1.70M & 29.17 / 0.8765 / 0.1801 \\
           & SR-STE (4:16) & 117.24 & 1.70M & 28.81 / 0.8684 / 0.1876\\
           & \cellcolor{orange!20}SLS (Ours) & \cellcolor{orange!20}116.31 & \cellcolor{orange!20}1.51M  & \cellcolor{orange!20}{\textbf{29.31}} / \cellcolor{orange!20}{\textbf{0.8800}} / \cellcolor{orange!20}{\textbf{0.1751}}\\
           \cline{2-5}
           & One-shot (2:8) & 117.24 & 1.70M & 29.11 / 0.8741 / 0.1812\\
           & SR-STE (2:8) & 117.24 & 1.70M & 28.75 / 0.8675 / 0.1898\\
           & \cellcolor{orange!20}SLS (Ours) & \cellcolor{orange!20}116.18 & \cellcolor{orange!20}1.49M  & \cellcolor{orange!20}{\textbf{29.20}} /  \cellcolor{orange!20}{\textbf{0.8767}} / \cellcolor{orange!20}{\textbf{0.1794}}\\
           \bottomrule[\heavyrulewidth]
       \end{tabular}
       \end{threeparttable}
   }
   \label{tab:deblurring_supp}
\end{table}
\begin{table}[t]
   \caption{Image super-resolution performace (PSNR$_\uparrow$) comparisons on benchmark datasets with the scaling factor of 4 and various $M$.
   } 
   \centering
   \resizebox{\linewidth}{!}{
       \begin{threeparttable}
       \begin{tabular}{clccc}
           \toprule[\heavyrulewidth]
                  Model & Method & GMACs & Num. Param.
                 & Set14 / B100 / Urban100\\
           \midrule[\heavyrulewidth] 
           \multirow{11}{*}{RFDN}& Unpruned & 39.86 & 828.75K & 28.52 / 27.51 / 25.91\\
           \cline{2-5}
           & Filter pruning & 10.44 & 214.77K & 16.50 / 17.32 / 15.52\\
           & One-shot (8:32) & 10.10 & 210.00K & 28.46 / 27.47 / 25.73\\
           & SR-STE (8:32) & 10.10 & 210.00K & 28.33 / 27.41 / 25.59\\
           & \cellcolor{orange!20}SLS (Ours) & \cellcolor{orange!20}10.05 & \cellcolor{orange!20}240.17K  & \cellcolor{orange!20}\textbf{28.50} / \cellcolor{orange!20}\textbf{27.50} / \cellcolor{orange!20}\textbf{25.84}\\
           \cline{2-5}
           & One-shot (4:16) & 10.10 & 210.00K & 28.44 / 27.46 / 25.73\\
           & SR-STE (4:16) & 10.10 & 210.00K & 28.35 / 27.42 / 25.60\\
           & \cellcolor{orange!20}SLS (Ours) & \cellcolor{orange!20}10.03 & \cellcolor{orange!20}250.66K  & \cellcolor{orange!20}{\textbf{28.50}} / \cellcolor{orange!20}{\textbf{27.49}} / \cellcolor{orange!20}{\textbf{25.82}}\\
           \cline{2-5}
           & One-shot (2:8) & 10.10 & 210.00K & 28.43 / 27.46 / 25.71\\
           & SR-STE (2:8) & 10.10 & 210.00K & 28.40 / 27.42 / 25.63\\
           & \cellcolor{orange!20}SLS (Ours) & \cellcolor{orange!20}10.02 & \cellcolor{orange!20}260.58K  & \cellcolor{orange!20}{\textbf{28.48}} /  \cellcolor{orange!20}{\textbf{27.49}} / \cellcolor{orange!20}{\textbf{25.82}}\\
           \bottomrule[\heavyrulewidth]
       \end{tabular}
       \end{threeparttable}
   }
   \label{tab:sr_supp}
\end{table}

\section{Adaptive Inference on Image Deblurring}
\label{supp_sec:supp_adaptive_inference}

\begin{figure}[!t]
    \newcommand{\wwp}{1\linewidth}
    \centering
    \includegraphics[width=\wwp]{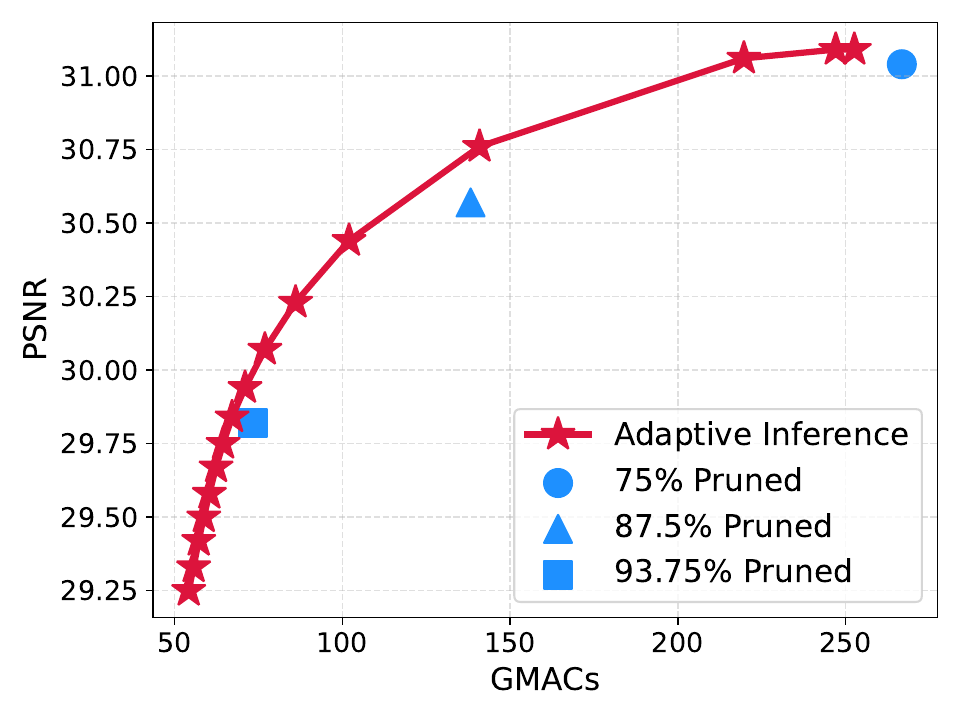}
    \caption{Results of the proposed adaptive inference method for image deblurring on GOPRO dataset.
    We use the identity function and 3 DMPHN models pruned by SLS with different target budgets.
    }
    \vspace{-1em}
    \label{fig:supp_adaptive_inference}
\end{figure}

In Section~\ref{sec:adaptive_inference} and Figure~\ref{fig:adaptive_inference_results} in the main manuscript, we showed that our adaptive inference exhibits an improved trade-off between PSNR and GMACs for super-resolution task.
We further validate the adaptive inference is also effective in image deblurring, as shown in Figure~\ref{fig:supp_adaptive_inference}.
Specifically, we trained an auxiliary network to predict the MSE error of the deblurred patches from 3 DMPHN models pruned by SLS and identity mapping (no processing).
Using the estimated MSE and the expected computational costs for the input patch, the adaptive inference path is determined by the Equation~\eqref{eq:adaptive_inference}.
For instance, a relatively sharp input patch can be fed to output by identity mapping or a model with high pruning ratio (\ie 93.75\%) and a severely blurry patch could be processed by the model with a higher accuracy.
The results show that our adaptive inference scheme can find the better trade-off between efficiency and the deblurring performance.
In all experimental results on the adaptive inference for both super-resolution and deblurring tasks, we report the computational costs containing the additional overhead from overlapping.

\section{Additional Implementation Details}
\label{supp_sec:details}

To show the effectiveness of our SLS framework, we performed additional experiments on state-of-the-art image deblurring and super-resolution networks.
For image deblurring, we used a residual UNet~\cite{nah2022clean}, SRN~\cite{Tao_2018_CVPR}, and DMPHN~\cite{Zhang_2019_CVPR_DMPHN}.
The residual UNet used in~\cite{nah2022clean} is a light-weight model in terms of GMACs while SRN and DMPHN are the relatively heavier models.
Following~\cite{nah2022clean}, we used a modified version of DMPHN by removing the multi-patch hierarchy for higher baseline accuracy.
All models for deblurring were trained and tested on GOPRO dataset~\cite{Nah_2017_CVPR}.
After pretraining each model for 2,000 epochs, we learn to prune the model for 2,000 additional epochs.

For image super-resolution, we used EDSR~\cite{lim2017enhanced}, CARN~\cite{ahn2018fast}, and RFDN~\cite{liu2020residual} models.
EDSR is a heavy architecture in terms of GMACs, CARN being a mid-weight, and RFDN is a light-weight architecture.
Specifically, we used EDSR-baseline which has 16 ResBlocks.
The super-resolution models were trained on DIV2K dataset~\cite{Agustsson_2017_CVPR_Workshops} and evaluated on 3 benchmark datasets, Set14~\cite{set14}, B100~\cite{martin_2001_ICCV}, and Urban100~\cite{Huang_CVPR_2015}.
The super-resolution models were pretrained for 300 epochs, followed by the pruning process for 300 additional epochs.

When applying SR-STE~\cite{zhou2021learning}, the model was trained from scratch for 4,000 epochs for deblurring, 600 epochs for super-resolution, respectively, for fair comparison.
We prune all convolutional layers with input channel dimension divisible by $M$.

For adaptive inference, the auxiliary convolutional network for MSE estimation consists of 0.15 M parameters and requires 3.45 GMACs for an HD image (1,280$\times$720).
For image deblurring, we crop the test images in GOPRO dataset into several patches with the size of 246$\times$266 and overlap 3 and 5 pixels in vertical and horizontal directions, respectively.
For image super-resolution, we crop the test images on Urban100 dataset into several patched with the size of 50$\times$50 with the stride of 48.
The restored patches are combined to the original image by removing the overlapping areas. 
We select the value of $\beta$ by a uniform sampling in the range of [0,10].

\subsection*{License of the Used Assets}

\begin{compactitem}[$\bullet$]
    \item GOPRO dataset~\cite{Nah_2017_CVPR} is a publicly available dataset released under CC BY 4.0 license.
    \item DIV2K dataset~\cite{Agustsson_2017_CVPR_Workshops} is made available for academic research purposes.
    \item B100 dataset~\cite{martin_2001_ICCV} is made available by Computer Vision Group, UC Berkeley
    \item Urban100 dataset~\cite{Huang_CVPR_2015} is made available at \url{https://github.com/jbhuang0604/SelfExSR}
\end{compactitem}

\subsection*{Limitations}
We study the layer-wise sparsity search technique on the general $N$:$M$ configurations for extremely efficient image restoration networks.
However, the latest Ampere-generation NVIDIA GPUs only support the acceleration of 2:4 sparsity pattern, limiting the acceleration of the efficient networks searched by SLS.
The efficient execution of our pruned models is our future work.

\clearpage
{\small
\bibliographystyle{ieee_fullname}
\bibliography{egbib}
}

\end{document}